# Improving Fitness Functions in Genetic Programming for Classification on Unbalanced Credit Card Datasets

Van Loi Cao, Nhien An Le Khac, Miguel Nicolau, Michael O'Neill, James McDermott

University College Dublin

**Abstract.** Credit card fraud detection based on machine learning has recently attracted considerable interest from the research community. One of the most important tasks in this area is the ability of classifiers to handle the imbalance in credit card data. In this scenario, classifiers tend to yield poor accuracy on the fraud class (minority class) despite realizing high overall accuracy. This is due to the influence of the majority class on traditional training criteria. In this paper, we aim to apply genetic programming to address this issue by adapting existing fitness functions. We examine two fitness functions from previous studies and develop two new fitness functions to evolve GP classifier with superior accuracy on the minority class and overall. Two UCI credit card datasets are used to evaluate the effectiveness of the proposed fitness functions. The results demonstrate that the proposed fitness functions augment GP classifiers, encouraging fitter solutions on both the minority and the majority classes.

## 1 Introduction

Credit cards have emerged as the preferred means of payment in response to continuous economic growth and rapid developments in information technology. The daily volume of credit card transactions reveals the shift from cash to card. Concurrently, the incidence of credit card fraud has risen with the loss of billions of dollars globally each year. Moreover, criminals have evolved increasingly more sophisticated strategies to exploit weaknesses in the security protocols. Therefore, there is great pressure on banks and financial organizations to improve their fraud detection models so that they may keep pace.

Classification models are generally applied to credit card fraud detection, but the problem domain poses serious challenges for such models. Various benchmark machine learning techniques are explored in the literature, Decision Tree Induction (C45), Artificial Neural Networks, Class Weighted Support Vector Machines, Bayes Networks, Artificial Immune Systems, Genetic Algorithms and so on [4, 7, 11]. However, the credit card fraud exhibits unique characteristics which render the task extremely challenging for any machine learning technique. The most common characteristic is that the credit card datasets are highly unbalanced, that is they admit and uneven distribution of class transactions. The fraud class is represented by only a small number of examples (minority class) while the legal class makes up the rest (majority class). The ratio of legal class size to fraud class size can vary up to hundred fold [11]. Using these datasets as training sets in the learning process can bias the learning algorithm resulting in poor accuracy on the minority class but high accuracy on the majority class [12]. This is because traditional training criteria such as overall success or error rate can be influenced by the larger number of instances from the majority class. Unfortunately, the minority class (fraud class) represents the main (positive) class, accurately classifying the fraud class is extremely important versus accurately classifying instances from the majority class. A small percentage of incorrectly classified examples from the minority class may lead to the loss of billion of dollars. Therefore, many methods for addressing unbalanced data have been developed to strengthen the classification algorithms for credit card fraud detection.

Generally speaking the approaches for handling unbalanced data can be categorized into two main groups. Firstly, in the external approach an artificially balanced distribution of class examples for training is created by sampling from the original unbalanced data sets. For instance, oversampling the minority class to boost representation, or editing of the majority class to decrease representation [1, 9]. Although these approaches can perform effectively, they have drawbacks. For instance, the training process can be burdened with a computational overhead from not only sampling algorithms, but also adjusted training data sets. In the internal approach the focus is to apply learning algorithms to the original unbalanced data sets[5, 6]. Interestingly, in Genetic Programming (GP), adapting the fitness function is known as a simple and efficient approach for addressing the imbalance in credit card fraud data. Solutions with good classification accuracy on minority class and overall are rewarded with better fitness whereas those that are only biased toward majority class are penalized with poor fitness [3, 2].

This paper aims to address the imbalance in credit card fraud data by applying a natural computing technique, namely Genetic Programming (GP). Our focus is on adapting the fitness function to evolve GP classifiers with significantly increased classification accuracy on the minority class while retaining high overall classification accuracy. The standard GP system is employed to investigate two fitness functions proposed by Bhowan [3], and based on the same two new fitness functions are presented. The experiments are conducted on two UCI standard datasets, the German Credit dataset, the Australian Credit Approval dataset, and the evaluation of these classifiers respects three basic measures, True Positive (TP), True Negative (TN) and overall accuracy. We also compare the results from GP classifiers evolved on these datasets to six benchmark machine learning methods including Support Vector Machines (c-SVM), two Bayesian Networks (BayesNet and NavieBayes), Decision Trees (J48) and two Artificial Neural Networks (Multilayer Perceptron and RBFNetwork).

The rest of this paper is organized as follows. In the next section, we give a short introduction to the genetic programming heuristic with emphasis on using

GP for classification. In Section 3, we briefly review two fitness functions from [3] and propose two new fitness functions. This is followed by a section detailing the experimental set up. Experimental results are presented with discussion in Section 5. The paper concludes with highlights and future directions.

## 2   GP for Classification

### 2.1   Genetic Programming

Genetic Programming was popularized by Koza in the 1990s [8]. It is an evolutionary paradigm that is inspired by biological evolution to find good solutions to a diverse spectrum of problems in the form of computer programs. The solutions are represented as variable shape and size parse trees, which allows the solutions to be evolved with respect to their structure and behavior. The programs which together constitute a population represent different candidates to the problem and combined with other programs to create new hopefully better solutions. This process is repeated over a number of generations until a good solution is evolved.

In terms of classification, GP has the ability to represent solutions to a wide range of complex problems since the representation is highly flexible. GP can leverage this flexibility in its representation to tackle classification problems that confound standard approaches including Decision Tree Induction, Statistical Classifiers and Neural Networks. Therefore, GP methods are well adapted to some classification problems than alternative classification methods [10].

### 2.2   Classification Strategy

For classification of credit card fraud data, the numeric (real) values returned by GP trees will be translated into class labels (fraud class and approval class) depending on whether the output of same falls within the non-negative or negative real number. That is, if the output of the GP classifier is greater or equal to zero, the transaction is said to belong to the minority class (fraud class), otherwise the transaction is predicted an instance of the majority class (approval class). Moreover, in order to achieve good accuracy on the minority class while retaining overall accuracy, the fitness function is adapted to evolve classifiers. Therefore, we examine two fitness functions from literature [3], and develop two new fitness functions. In this paper, we use the term GPout as representing the output from the GP tree.

## 3   Improving Fitness Function

In this section, we will investigate two fitness functions that were proposed by Bhowan [3] ($f_{equal}$ and $f_{errors}$). $f_{equal}$ is designed to treat the classification accuracy of both classes as equally important. $f_{errors}$ is given by $f_{equal}$ with the error on both classes. Thus $f_{errors}$ is sensitive to not only the minority

44     Authors Suppressed Due to Excessive Length

class accuracy but also the classification ability. With the aim of raising the fraud class detection accuracy while retaining considerable overall classification accuracy, two new fitness functions are presented as modification of $f_{errors}$.

### 3.1 Previous Fitness Function

**Fitness Function 1.** Eq. 1 was presented in [3]. The classification accuracy of both classes is treated as equally important. Two aspects of this fitness function are studied in this paper. Firstly, we examine the effectiveness of balancing the TP and TN rates by treating as equally important on both classes. Secondly, the basic framework for evaluating the proposed fitness functions is described.

$$f_{equal} = \frac{TP}{TP+FN} + \frac{TN}{TN+FP} \quad (1)$$

**Fitness Function 2.** The fitness function $f_{errors}$ proposed in [3] was formed by combining the fitness function in Eq. 1 with the error rate of each class, given by Eq. 3. The error function aims to differentiate between solutions which score the same classification accuracy on each class but with different internal classification models. Solutions with smaller levels of error for each class are closer to correctly labeling any incorrectly predicted examples. These solutions will have better classification models, and are favored over solutions with a larger levels of error [3].

$$f_{errors} = \frac{TP}{TP+FN} + \frac{TN}{TN+FP} + (1-Err_{min}) + (1-Err_{maj}) \quad (2)$$

Where,

$$Err_c = \frac{(|P_c^{mx}| + |P_c^{mn}|)}{2}. \quad (3)$$

The error function for class c, $Err_c$, in Eq. 3, is estimated using the largest and smallest incorrect GPout values for a particular class, $P_c^{mx}$ and $P_c^{mn}$ respectively. The absolute value for incorrect GPout is taken because it is negative on the minority class. These values are scaled to between 1 and 0 where 1 indicates the highest level of error and 0 the lowest.

### 3.2 Proposed Fitness Function

**Proposed Fitness Function 1.** Eq. 3 illustrates that $Err_c$ is computed based on the largest and smallest incorrect GPout values on a particular class c. Therefore, if the value of $|P_c^{mn}|$ is extremely large, the level of error for class c will be very large even though there are few incorrect GPout values. This will give weight to the classification accuracy of the class with the extreme incorrect GPout value. Therefore, the fitness function $f_{errors}$ is more sensitive to extreme incorrect GPout values.

In order to address this issue, $Err_c$ in Eq. 3 is modified by using the mean of the all incorrect GPout values on particular class to estimate the level of error for the class, namely $Err_c^{mean}$. Where, $P_{ci}$ and $m$ are the $i-th$ incorrect GPout value on class $c$ and the number of incorrect classification examples on class $c$ respectively. By doing this, the fitness function $f_{errors\ mean}$ is less sensitive to the extreme incorrect GPout values than $f_{errors}$.

$$f_{errors\ mean} = \frac{TP}{TP+FN} + \frac{TN}{TN+FP} + (1-Err_{min}^{mean}) + (1-Err_{maj}^{mean}) \quad (4)$$

Where,

$$Err_c^{mean} = \frac{\sum_{i=1}^{m}(|P_{ci}|)}{m}. \quad (5)$$

**Proposed Fitness Function 2.** When the distribution of the incorrect GPout values is skewed, the fitness function $f_{errors\ mean}$ may be still sensitive to the extreme values. In this case, the error function $Err_c^{median}$ is proposed by using the median of the incorrect GPout values on particular class to estimate the error. Where, $P_{ci}$ and $m$ are the $i-th$ incorrect GPout value on class $c$ and the number of incorrect classification examples on class $c$ respectively. Eq. 6 shows that the fitness function $f_{errors\ median}$ is partly contributed by $Err_c^{median}$, which is not sensitive to the extreme incorrect values and the skewed distribution of the incorrect values.

$$f_{errors\ median} = \frac{TP}{TP+FN} + \frac{TN}{TN+FP} + (1-Err_{min}^{median}) + (1-Err_{maj}^{median}). \quad (6)$$

Where,

$$Err_c^{medium} = \begin{cases} P_{c([\frac{m}{2}]+1)} : m = 2k+1, k \in N, \\ \frac{P_{c([\frac{m}{2}])}+P_{c([\frac{m}{2}]+1)}}{2} : m = 2k, k \in N. \end{cases} \quad (7)$$

## 4 Experimental Settings

### 4.1 Credit Card Datasets

Two credit card datasets that were previous used classification methods are also introduced. The German credit dataset which contains categorical attributes was first procured by Prof. Hofmann. It was then converted into the numeric version by Strathclyde University. Each transaction is described by 24 attributes and one class attribute which classifies the transaction into the good or bad class. There are 700 instances for the good class and 300 instances for the bad class. For the German credit card dataset, we randomly select 50 percent on each class for the training set, other 50 percent on each class for the testing set in order to remain the imbalance between the good and bad class in the two sets.

The Australia Credit Approval dataset originated from Quinlan. All attribute names and values of the data had been changed to meaningless symbols for the confidentiality reasons. There are 690 instances each of which belongs to approval class or risk class (383 approval, 307 risk instances respectively). Each instance is described by 14 real-valued attributes and one class attribute. However, the data is slightly unbalanced, so we randomly sample only 30 percent for the training set, and 30 percent for the testing set on the risk class. Furthermore, we retain 50 percent for the training set and the testing set on the approval class. In order to weight all attributes equally, a Min-max normalization technique is employed to map the values of all attributes of the two datasets to the interval [0.0, 1.0].

### 4.2 Evolutionary Parameters

A set of parameter sweep experiments were carried out to search for the best evolutionary parameter settings for the credit card fraud data. Five different settings for $f_{equal}$ were tested on two datasets, with successively increasing population sizes, as illustrated in Figs. 1(a) and 1(b). The average fitness values are computed with respect to the different population sizes (100, 200, 300, 400 and 500) while keeping the number of generation at 1000 over 30 runs. As presented in Figs. 1(a) and 1(b), the parameters of Pop = 500 and Gen = 1000 produce the smallest fitness values on the both data sets over 1000 generations. Therefore, we use the settings for all experiments in this paper. Moreover, our experiments employed the standard GP system written on Java programming language with the standard parameter settings: crossover = 0.9, mutation = 0.1 and tournament size = 3. The evolutionary parameter settings are listed in Table 1.

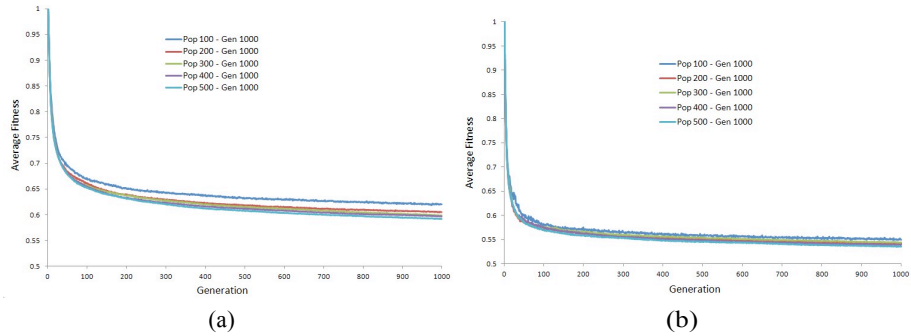

**Fig. 1.** Average fitness values across 30 runs for a range of parameter settings on the two datasets. (a) the German Credit dataset. (b) the Australia Credit Approval dataset.

### 4.3 Experiments

This section presents the settings for the experiments that were conducted in this study. Four sets of experiments are conducted in order to estimate the utility of the proposed fitness functions. The first aims to uncover how important balance between TP and TN rates are when the minority and majority classes are treated with equal importance. The result will be a basis on which to evaluate the classifiers in the following experiments. The second experiment is conducted to investigate how the error function, Eq. 3, of the fitness function $f_{errors}$ effects the TP rate and classification ability. Finally, the third and fourth experiments will examine the extent to which the proposed fitness functions achieve high overall classification accuracy as the TP rate increases.

**Table 1.** Evolutionary parameter settings

| Parameter | Value |
|---|---|
| Number of run | 30 |
| Population size | 500 |
| Number of generation | 1000 |
| Crossover Probability ($P_{cross}$) | 0.9 |
| Mutation Probability ($P_{mut}$) | 0.1 |
| Selection | Tournament |
| Tournament size | 3 |

The effectiveness of the proposed fitness functions are also benchmarked against six well-known machine learning techniques. These techniques were trained and tested on the above data sets. The tested learning techniques include Support Vector Machines (c-SVM), Bayes Networks (BayesNet and NaivieBayes), Decision Trees (J48) and Artificial Neural Networks (Multilayer Perception and RBF networks). The results from applying the proceeding learning methods to the above datasets are delineated in the following section.

## 5 Result and discussion

This section presents the experimental results realized by applying GP to address the imbalance in credit card dataset. The experiments are implemented on the standard GP system with the evolutionary parameters reported in Table 1. The performance of each GP classifier is evaluated along three dimensions three, TP, TN rates and overall accuracy. Moreover, the implementations of the six benchmark techniques are performed in Weka. The results are shown in the tables 2 and 3.

Table 2 and 3 illustrate the results of applying GP algorithm with four different fitness functions, and six benchmark machine techniques on German Credit and Australian Credit Approval data sets. The first subtable reports the results of various tested learning techniques, namely Support Vector Machine (c-SVM), two Bayesian Networks (NaiveBayes and BayesNet), Decision Tree (J48) and two Artificial Neural Networks (Multilayer Perceptron and RBFNetwork). The results of genetic programming classifiers with the four fitness functions ($f_{equal}$, $f_{errors}$, $f_{error\ mean}$ and $f_{errors\ median}$) are presented in the last rows of these tables. The term $C_i$ represents the GP classifier with its fitness function of $f_i$.

Observe that the fitness function $f_{errors\ mean}$ improve the GP classifier's performance on both datasets significantly. The classifier not only achieve increases in TP rate but also sustained good overall accuracy (0.76, 0.848 in Table 2, and 0.84, 0.864 Table 3) in comparison to the GP classifiers $C_{equal}$ and $C_{errors\ median}$ (0.757, 0.757 in Table 2, and 0.803, 0.877 in Table 3). Although TP rates of the classifier $C_{errors\ mean}$ are slightly lower than those of the classifier $C_{errors}$, the balance between TP rate and overall accuracy of $C_{errors\ mean}$ is better than those of $C_{errors}$.

**Table 2.** The true positive, true negative and accuracy rates of learning methods on the German Credit dataset.

| Technique | TP Rate | TN Rate | Accuracy |
|---|---|---|---|
| c-SVM$_{(gamma=1, weights=0.45)}$ | **0.773** | **0.823** | **0.808** |
| NaiveBayes | 0.533 | 0.840 | 0.748 |
| BayesNet | 0.380 | 0.866 | 0.720 |
| J48$_{(reducederrorprunning=T, numfolds=3)}$ | 0.507 | 0.866 | 0.758 |
| RBFNetwork$_{(Seed=76, numcluster=9)}$ | 0.587 | 0.826 | 0.754 |
| MultilayerPerceptron | **0.760** | **0.886** | **0.848** |
| $f_{equal}$ | 0.757 | 0.758 | 0.757 |
| $f_{errors}$ | 0.823 | 0.616 | 0.678 |
| $mean\ f_{errors}$ | **0.770** | **0.745** | **0.752** |
| $median\ f_{errors}$ | 0.743 | 0.751 | 0.749 |

Interestingly, these tables reveal that GP performs better on the unbalanced credit card data sets than the majority of the benchmark techniques. The TP rates of the GP classifier $C_{errors\ mean}$ (0.77 in Table 2 and 0.840 in Table 3) are significantly higher than those of several benchmark techniques (0.533(NaiveBayes), 0.380(BayesNet) or 0.507(J48) in Table 2, and 0.600(NaiveBayes), 0.770(BayesNet) or 0.710(RBENetwork) in Table 3). However, c-SVM whose a weighted parameter produced the best results of TP rate and overall accuracy, 0.773 and 0.808 on the first data set, and 0.900 and 0.846 on the second data set respectively. Overall, the results in this section demonstrates that GP can overcome the imbalance in credit card data, and judicious selection of fitness functions can make the search more efficient.

## 6  Conclusion

In this paper, we address the problem that practitioners often face when classifying credit card datasets. The problem is the imbalance between the approval class (majority class) and the fraud class (minority class), which classifiers tend to yield poor accuracy on the fraud class despite of high overall accuracy. In this paper, we propose two fitness functions to evolve GP classifiers with significantly increased classification accuracy on the minority class while retaining high overall classification accuracy. The GP classifiers are tested using the two UCI credit card datasets, and the classification ability of them are also compared against six well-known learning techniques. The experimental results demonstrate that GP can handle the imbalance in credit card datasets, and judicious selection of fitness functions can make the search more efficient.

**Table 3.** The true positive, true negative and accuracy rates of learning methods on the Australia Credit Approval dataset.

| Technique | TP Rate | TN Rate | Accuracy |
|---|---|---|---|
| c-SVM$_{(gamma=1, weights=0.45)}$ | **0.900** | 0.818 | 0.846 |
| NaiveBayes | 0.600 | 0.943 | 0.825 |
| BayesNet J48$_{(reducederrorprunning=T, numfolds=3)}$ | 0.770 | 0.917 | 0.866 |
|  | **0.840** | **0.885** | **0.870** |
| RBFNetwork$_{(Seed=76, numcluster=9)}$ | 0.710 | 0.932 | 0.856 |
| MultilayerPerceptron | 0.780 | 0.932 | 0.880 |
| $f_{equal}$ | 0.803 | 0.915 | 0.877 |
| $f_{errors}f_{errors}$ | 0.842 | 0.853 | 0.849 |
| $mean f_{errors}$ | **0.840** | **0.877** | **0.864** |
| median | 0.803 | 0.914 | 0.876 |

In terms of future work we plan to use the AUC (the Area Under Curve) to evaluate the GP classifiers with proposed fitness functions.

# 7  Acknowledgements

This work is funded by Vietnam International Education Development (Vied), and supported by Prof Michael ONeill.